\documentclass[]{gensi}
\usepackage{tabularx}
\usepackage[toc,page,header]{appendix}

\usepackage{minitoc}
\usepackage{cleveref} 
\usepackage{microtype}
\usepackage{subcaption}
\usepackage{amsmath} 
\usepackage{amssymb} 
\usepackage{amsfonts}       
\usepackage{pgfplots}
\usepackage{pgfplotstable}
\usepackage[dvipsnames]{xcolor}
\usepackage{CJKutf8}
\usetikzlibrary{patterns}
\usepackage{multirow}
\usepackage{setspace}
\usepackage{xcolor}
\usepackage{hyperref}
\tcbuselibrary{theorems}
\usepackage{float}
\usepackage{caption}
\usepackage{wrapfig}
\usepackage{xspace}
\usepackage{makecell}
\usepackage{graphicx}      
\usepackage{subcaption}    
\usepackage{amsthm}
\usepackage{marvosym}

\usepackage{bbding}
\newtheorem{definition}{Definition}
\newtheorem{corollary}{Corollary}

\newlength{\MCwidth}
\newcommand{\method}{\textsc{FLEX}}

\usepackage{tcolorbox}
\newtcolorbox{DefinitionBox}{
  colback=blue!5,
  colframe=blue!80,
  boxrule=0.5pt,
  arc=2pt,
  left=2pt,
  right=2pt,
  top=2pt,
  bottom=2pt,
}

\newtcolorbox{CorollaryBox}{
  colback=gray!5,
  colframe=gray!80,
  boxrule=0.5pt,
  arc=2pt,
  left=2pt,
  right=2pt,
  top=2pt,
  bottom=2pt,
}

\usepackage{fix-cm}

\usepackage{natbib}
\usepackage{latexsym}

\usepackage{url}
\usepackage{amssymb}
\usepackage[utf8]{inputenc}
\usepackage{microtype}
\usepackage{booktabs}
\usepackage{pifont} 
\usepackage{multirow}
\usepackage{makecell}
\usepackage{paralist}
\usepackage{xspace}
\usepackage{color}
\usepackage{xcolor}
\usepackage{colortbl}
\usepackage{adjustbox}
\usepackage{hyperref} 
\usepackage[edges]{forest}
\usepackage{tikz} 
\usepackage{caption}
\usepackage{amsfonts}
\usepackage{tcolorbox}
\usepackage{algorithm}
\usepackage{algpseudocode}
\usepackage{amsthm}

\usepackage{mathtools}
\usepackage[version=4]{mhchem}
\usepackage{array}  
\usepackage{graphicx}

\hypersetup{
    colorlinks,
    linkcolor={blue!80!black},
    citecolor={blue!80!black},
}
\tikzset{
    root/.style =             {align=center, text width=1cm, rounded corners=3pt, line width=0.3mm, fill=gray!10, draw=gray!80, font=\small},
    demographic/.style =         {align=center, text width=1.8cm, rounded corners=3pt, line width=0.3mm, fill=blue!10, draw=blue!80, font=\footnotesize},
    demographic_work/.style =    {align=center, text width=10cm, rounded corners=3pt, line width=0.3mm, fill=blue!10, draw=blue!0, font=\footnotesize},
    character/.style =         {align=center, text width=1.8cm, rounded corners=3pt, line width=0.3mm, fill=red!10, draw=red!80, font=\footnotesize},
    character_work/.style =    {align=center, text width=10cm, rounded corners=3pt, line width=0.3mm, fill=red!10, draw=red!0, font=\footnotesize},
    personalization/.style =           {align=center, text width=1.8cm, rounded corners=3pt, line width=0.3mm, fill=cyan!10, draw=cyan!80, font=\footnotesize},
    personalization_work/.style =      {align=center, text width=10cm, rounded corners=3pt, line width=0.3mm, fill=cyan!10, draw=cyan!0, font=\footnotesize},
    risk/.style =         {align=center, text width=1.8cm, rounded corners=3pt, line width=0.3mm, fill=orange!10, draw=orange!80, font=\footnotesize},
    risk_work/.style =    {align=center, text width=10cm, rounded corners=3pt, line width=0.3mm, fill=orange!10, draw=orange!0, font=\footnotesize},
}

\newcommand{\cmark}{\ding{51}}
\newcommand{\xmark}{\ding{55}}

\usepackage{CJK}

\usepackage{amsmath,amsfonts,bm}

\def\eqref#1{equation~\ref{#1}}

\def\1{\bm{1}}

\DeclareMathAlphabet{\mathsfit}{\encodingdefault}{\sfdefault}{m}{sl}
\SetMathAlphabet{\mathsfit}{bold}{\encodingdefault}{\sfdefault}{bx}{n}

\title{
    \method: Continuous Agent Evolution via \\ Forward Learning from Experience
}

\author[1,6,*]{Zhicheng Cai}
\author[1,4,*]{Xinyuan Guo}
\author[1*]{Yu Pei}
\author[1,3,\dagger]{Jiangtao Feng}
\author[3,5]{Jinsong Su}
\author[2,6]{\\Jiangjie Chen}
\author[1,6]{Ya-Qin Zhang}
\author[1,6]{Wei-Ying Ma}
\author[2,6]{Mingxuan Wang}
\author[1,6,\ddagger]{Hao Zhou}

\affiliationLast[1]{Institute for AI Industry Research (AIR), Tsinghua University}
\affiliationSameLine[2]{ByteDance Seed}
\affiliationNewLine[3]{Shanghai Artificial Intelligence Laboratory}
\affiliationSameLine[4]{University of Chinese Academy of Sciences}
\affiliationNewLine[5]{School of Informatics, Xiamen University} 
\affiliationSameLine[6]{SIA-Lab of Tsinghua AIR and ByteDance Seed}

\contribution[*]{Equal Contribution}
\contribution[\dagger]{Project Lead}
\contribution[\ddagger]{Corresponding Author\\}
\contributionNewLine[]{ Correspondence: \href{mailto:zhouhao@air.tsinghua.edu.cn}{zhouhao@air.tsinghua.edu.cn}}

\abstract{

Autonomous agents driven by Large Language Models (LLMs) have revolutionized reasoning and problem-solving but remain static after training, unable to grow with experience as intelligent beings do during deployment.
We introduce \textbf{F}orward \textbf{L}earning with \textbf{EX}perience~(\method{}), a gradient-free learning paradigm that enables LLM agents to continuously evolve through accumulated experience.
Specifically, \method{} cultivates scalable and inheritable evolution by constructing a structured experience library through continual reflection on successes and failures during interaction with the environment.
\method{} delivers substantial improvements on mathematical reasoning, chemical retrosynthesis, and protein fitness prediction (up to 23\% on AIME25, 10\% on USPTO50k, and 14\% on ProteinGym).
We further identify a clear scaling law of experiential growth and the phenomenon of experience inheritance across agents, marking a step toward scalable and inheritable continuous agent evolution.
Project Page: \url{https://flex-gensi-thuair.github.io}.
}

\begin{document}
    
    \maketitle
    \vspace{-7em}
    \begin{center}
        \includegraphics[width=0.9\linewidth]{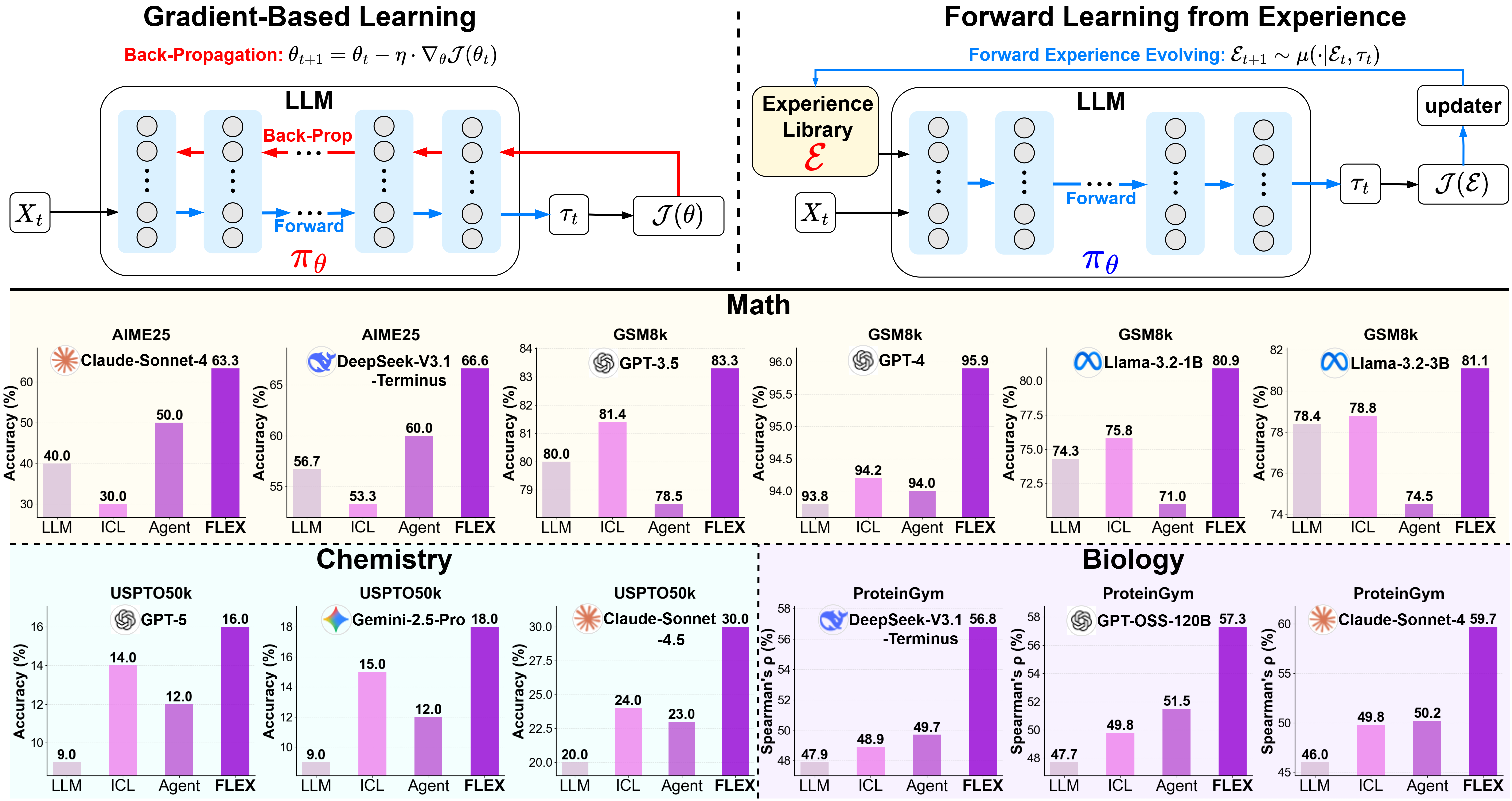}
        \vspace{-0.5em}
        \captionof{figure}{An overview of our FLEX paradigm and main results. \textbf{Top:} A comparison between traditional gradient-based learning, which uses \textbf{Back}-Propagation as the optimizing method and our proposed \textbf{Forward} Learning from Experience paradigm. \textbf{Bottom:} Main experimental results demonstrating FLEX's effectiveness. We evaluate FLEX against strong baselines across three challenging scientific domains (Mathematics, Chemistry, and Biology) on a diverse suite of over 10 models, where FLEX consistently and substantially outperforms the baselines with the cost \textit{less than 100\$} for both training and evaluation of a single agent.}
        \label{fig:main_method_overview}
    \end{center}

    \newpage
    \tableofcontents
    \newpage

    \section{Introduction}
Autonomous agents have achieved remarkable advances across diverse domains, including code generation (e.g., Claude Code~\cite{anthropic2025claude}), scientific discovery (e.g., Biomni~\cite{huang2025biomni}, STELLA~\cite{jin2025stellaselfevolvingllmagent}, and SciMaster~\cite{chai2025scimastergeneralpurposescientificai}), and in-depth research (e.g., Tongyi, OpenAI, and Gemini DeepResearch~\cite{tongyidr,openai2025deepresearch,google2025gemini}), showcasing their immense potential to solve complex, open-ended real-world tasks end to end. 

However, pretrained agents remain static with frozen parameters, fundamentally precludes learning from on-the-fly experience concluded during trial-and-error, causing substantial performance degradation on challenging or untrained tasks~\cite{wu2023autogen,liu2023agentbench}.
While gradient-based learning has underpinned model optimization~\cite{rumelhart1986learning,ruder2016overview,hu2022lora}, it is inherently ill-suited for continuous agent evolution due to three key obstacles:
(i) back-propagation incurs prohibitive computational costs~\cite{goodfellow2013empirical,cai2025training};
(ii) parameter-centric paradigm suffers from catastrophic forgetting~\cite{abbas2023loss,hadi2023large}, failing to effectively learn form accumulated experiences over times;
(iii) the closed-source nature of most state-of-the-art LLMs~\cite{google2025gemini2_5_pro,openai2025gpt5,anthropic2025sonnet4.5,anthropic2025claude4,xai2025grok4} renders direct parameter optimization infeasible.

Recent self-evolving paradigms attempt to overcome this limitation by evolving \textit{non-parametric} components (\textit{i.e.,} prompts~\cite{yuksekgonul2024textgrad,zhang2025agentic}, workflows~\cite{lin2025se,zhang2025multi,shang2024agentsquare}, and tools~\cite{qian2023creator,qiu2025alita}) through environmental feedback and textual gradient~\cite{yuksekgonul2024textgrad,shinn2023reflexion,zhang2024revolve}. 
Nevertheless, these approaches encounter three vital bottlenecks towards continuous agent evolution: 
(i) their evolving components are task-specific, thus they cannot generalize across tasks, limiting cross-task evolution.
(ii) Their evolving components (\textit{i.e.} prompts, workflows, and tools) are not continuously scalable, thus they can only leverage a limited set of experiences, preventing performance from scaling with accumulated knowledge.
(iii) Their evolving components are model-specific, thus new agents must interact from scratch during deployment, unable to continuously evolve from agents' previous off-policy experiences, leading to substantial computational overhead.

To address these limitations, we propose \textbf{F}orward \textbf{L}earning from \textbf{Ex}perience~(\textbf{\method{}}), a novel learning paradigm that shifts learning from modifying \textit{model parameters} to constructing and leveraging an evolvable \textit{experience library} (Fig.\ref{fig:main_method_overview}).
In contrast to traditional gradient-based learning paradigms, which optimize through both forward- and backward-propagation, \method{} involves mere forward passes in three stages: 
(i) exploring forwardly to acquire extensive problem-solving experiences;
(ii) evolving the experience library via the updater;
(iii) guiding forward reasoning by utilizing pertinent experiences from the library.

Hence, \method{} overcomes the prevailing bottlenecks with three profound properties for continuous evolution: 
(i) \method{} evolves a persistent experience library that continuously aggregates cross-task experiences, ensuring that knowledge is retained and available for future use.
(ii) The library continuously expands and refines, enabling agents to progressively acquire deeper insights and knowledge. This accumulation allows agent performance to scale with experience.
(iii) \method{} stores strategies distilled from experiences in a semantic form, making the library both transferable and interpretable. This design enables seamless inheritance across agents, avoiding redundant and costly re-training.

To validate the general effectiveness of \method{} paradigm, we conduct extensive experiments across diverse challenging scientific domains, including Olympiad-level mathematics (AIME25~\cite{aime2025problems}), chemical retrosynthesis (USPTO50k~\cite{schneider2016s}), and protein fitness prediction (ProteinGym~\cite{notin2023proteingym}). \method{} demonstrates substantial and consistent improvements on these tasks, from $40\%$ to $63\%$ on AIME25, $20\%$ to $30\%$ on USPTO50k, and $46\%$ to $60\%$ on ProteinGym, exhibiting great enhancement in the capacity of reasoning and knowledge leverage. 

In summary, our contributions are as follows:
\begin{itemize}
\item We propose \textbf{\method{}}, a new learning paradigm for the agentic era. It redefines learning as a forward exploration and experience distillation process, enabling LLM agents to continuously evolve through accumulated experience without gradient-based tuning.
\item We provide a comprehensive framework for \method{}, including a unified mathematical formulation with theoretical justifications, a practical instantiation with concrete mechanisms, and an empirical demonstration of its effectiveness on diverse scientific benchmarks.
\item We identify and empirically validate a \textit{scaling law} of the experience library, showing that agent performance scales predictably with accumulated experience. 
We also introduce and demonstrate the principle of \textit{experience inheritance}, where distilled experience can be transferred between agents in a plug-and-play manner, enabling instant knowledge assimilation and bypassing redundant learning.
\end{itemize}

\section{Forward Learning with Experience}
In this section, we establish a unified mathematical formulation of \method{} paradigm. 
A learning paradigm typically consists of two key components: an \emph{optimization objective} (defining \textit{what} to learn) and an \emph{optimization process} (defining \textit{how} to learn it). We next illustrate how \method{} implements these components.

\subsection{Overview of \method{}}
\method{} enables LLM agents to continually evolve through \textbf{learning from experience}. 
Instead of fine-tuning model parameters, \method{} focuses on constructing an \textbf{experience library} that captures and reuses knowledge distilled from past problem-solving trajectories, guiding future reasoning.

\method{} paradigm introduces a forward learning loop with two interleaved phases:
(i) extensive \textit{forward} exploration by the actor to collect and distill new experiences, which are integrated into the library by the updater; 
(ii) utilizing the most pertinent experiences to guide \textit{forward} reasoning on new tasks. 
As the experience library expands, it accumulates diverse, high-quality problem-solving strategies, allowing agents to leverage prior knowledge for increasingly superior reasoning, all without gradient-based parameter updates. 
Through this iterative process, agents continuously learn from accumulated experience, progressively enhancing their performance on challenging problems.
Fig.~\ref{fig:main_method_overview} illustrates the overall workflow of \method{}, showing the interaction between the actor $\pi$, updater $\mu$, and the experience library $\mathcal{E}$.

Specifically, the LLM agent $\pi$ remains frozen while being driven to explore the environment extensively, generating interaction trajectories $\{\tau \mid X, \pi\}$ for a given query $X$.
These trajectories are dynamically consolidated into the ever-evolving experience library $\mathcal{E}$ by an auxiliary updater agent $\mu$. 
Thus the library update can be denoted as $\mathcal{E}_{new}\sim\mu(\cdot\mid\mathcal{E}_{old},\{\tau\mid X,\pi\})$. The learned library accumulates distilled knowledge over time and subsequently serves as a guidance source for reasoning on future tasks.

At inference time, the most pertinent and helpful experiences are identified and fetched through $\varepsilon\sim\rho(\cdot\mid q,\mathcal{E})$ from the learned experience library $\mathcal{E}$ given query $q$.  
Conditioned on these retrieved experiences $\varepsilon$, $\pi$ is enabled to invoke learned strategies and latent cognitive patterns, producing superior response $\pi(\cdot \mid q, \varepsilon)$ that exhibit significantly enhanced reasoning capability and expert-level knowledge utilization compared to the base model.

Formally, we define the optimization objective of \method{} as constructing an optimal experience library $\mathcal{E^*}$ that maximizes expected correctness on training tasks, and operationalize it through a forward update process.
\paragraph{Optimization Objective.}
The optimization goal of \method{} is to construct an experience library that maximizes the expected correctness of the model’s outputs, thus we have:
\begin{definition}[Optimization Objective of \method{}]
\label{def:objective}
Given a training set $\mathcal{D} = \{(X_i, Y_i)\}_{i=1}^N$, where $X_i$ is a query and $Y_i$ is the corresponding ground truth, the objective function $\mathcal{J}(\mathcal{E})$ and the optimal experience library $\mathcal{E}^*$ is defined as:
\begin{equation}
\begin{aligned}
\mathcal{J}(\mathcal{E}) 
&= \mathbb{E}_{(X_i, Y_i)\sim\mathcal{D},\, \varepsilon_i\sim\rho(\cdot\mid X_i,\mathcal{E})}
  \big[\Phi\big(\pi(\cdot\mid X_i,\varepsilon_i),\, Y_i\big)\big], \\
\mathcal{E}^* 
&= \arg\max_{\mathcal{E}}\, \mathcal{J}(\mathcal{E})
\end{aligned}
\label{eq:obj}
\end{equation}
where $\rho(\cdot\mid X_i,\mathcal{E})$ retrieves experiences $\varepsilon_i$ from library $\mathcal{E}$ given query $X_i$, $\pi(\cdot\mid X_i,\varepsilon_i)$ is the LLM agent conditioned on both the query and retrieved experiences, and $\Phi(\hat{Y}, Y_i)$ measures the correctness of prediction $\hat{Y}$ against ground truth $Y_i$.
\end{definition}
This objective serves as the guiding principle for the experience library evolution in \method{}, explicitly linking the quality of the experience library to the model's ultimate performance on target tasks.

\paragraph{Optimization Process.}
\method{} operationalizes the objective in Definition~\ref{def:objective} through a forward learning process. 
Instead of relying on gradient-based parameter updates, learning occurs via forward probabilistic updates to the experience library, carried out by the updater agent $\mu$.
This process distills knowledge from interaction trajectories generated by $\pi$, steering the experience library toward optimization of the objective function. 
Therefore, the learning update rule for \method{} can be analogously defined as follows:

\begin{definition}[Update Rule of \method{}]
\label{def:forward_update}
The learning process in \method{} is governed by a forward update rule that evolves the experience library through forward probabilistic updates with experiential exploration, defined as:
\begin{equation}
\begin{aligned}
\mathcal{E}_{i+1} 
&\sim \mu\big(\cdot \,\mid\, \mathcal{E}_i, \{\tau_i \mid X_i, \pi\}\big), \\
\nabla_{\mathcal{E}}\mathcal{J}(\mathcal{E}_i) 
&\triangleq \mu\big(\cdot \,\mid\, \mathcal{E}_i, \{\tau_i \mid X_i, \pi\}\big) - \mathcal{E}_i
\end{aligned}
\label{eq:meta_update}
\end{equation}
where $\{\tau_i\mid X_i,\pi\}$ represents the set of interaction trajectories generated by actor agent $\pi$ given input $X_i$, $\mu(\cdot\mid\mathcal{E}_{i},\{\tau_i\mid X_i,\pi\})$ is the update distribution that distills new experiences from $\{\tau_i\mid X_i,\pi\}$ into $\mathcal{E}_i$, and $\nabla_{\mathcal{E}}\mathcal{J}(\mathcal{E})$ denotes the optimization direction for improving the experience library.
\end{definition}

In contrast to traditional gradient-based methods, which involves computing gradients of the objective function with respect to model parameters and updating them via backpropagation, i.e., $\theta_{i+1} = \theta_i - \nabla_\theta \mathcal{J}(\theta_i)$, \method{} introduces a fundamentally distinct paradigm by updating the experience library $\mathcal{E}$ through \textit{forward probabilistic exploration} rather than optimizing parameters via \textit{backward gradient propagation}, allowing the experience library to dynamically adapt and enhance model performance through accumulated experiential knowledge.
Fig.~\ref{fig:main_method_overview} exhibits the differences between gradient-based learning and \method{}.

\subsection{Information-Theoretic Insights into \method{}}
To gain deeper insight into how the experience library drives learning, we reinterpret the optimization objective from an information-theoretic perspective. 
This reformulation reveals that \method{} effectively learns to reduce predictive uncertainty and maximize the information gained from retrieved experiences.

Typically, $\Phi(\hat{Y_i},Y_i) = \mathbb{I}(\hat{Y_i}\!=\!Y_i)$, thus the objective $\mathcal{J}(\mathcal{E})$ in Eq.~\ref{eq:obj} can be equivalently written as maximizing the log-likelihood:
\begin{equation}
\begin{aligned}
\mathcal{J}(\mathcal{E}) 
&= \mathbb{E}_{(X_i,Y_i)\sim\mathcal{D},\ \varepsilon_i\sim\rho(\cdot\mid X_i,\mathcal{E})}
\left[\log \pi(Y_i\mid X_i,\varepsilon_i)\right], \\[3pt]
\mathcal{E}^* 
&= \arg\max_{\mathcal{E}}\mathcal{J}(\mathcal{E})
\end{aligned}
\label{eq:objective}
\end{equation}
Further, the log-likelihood can be equivalently expressed in terms of conditional entropy $\mathcal{H}$ and KL divergence:
\begin{equation}
\mathcal{J}(\mathcal{E}) = 
-\mathbb{E}_{\substack{(X_i,Y_i)\sim\mathcal{D} \\ \varepsilon_i\sim\rho(\cdot\mid X_i,\mathcal{E})}}
\Big[\mathcal{H}(Y_i\mid X_i,\varepsilon_i) + 
\mathrm{KL}\big(p^*(\cdot\mid X_i,\varepsilon_i)\,\|\,\pi(\cdot\mid X_i,\varepsilon_i)\big)\Big]
\end{equation}
After the extensive pre- and post-training, the LLM $\pi$ is assumed to possess sufficient comprehension and reasoning capabilities to effectively utilize input information. 
Hence, the model distribution $\pi(\cdot\mid X_i,\varepsilon_i)$ can be regarded as sufficiently close to the true conditional distribution $p^*(\cdot\mid X_i,\varepsilon_i)$. 
Consequently, the second term becomes negligible, and the optimization objective can be approximated as minimizing the expected conditional entropy:

\begin{corollary}[Information-Theoretic Reformulation of the Objective]
\label{cor:obj}
The optimal experience library $\mathcal{E}^*$ that maximizes $\mathcal{J}(\mathcal{E})$ can be approximated by minimizing the expected conditional entropy:
\begin{equation}
\mathcal{E}^* =
\arg\max_{\mathcal{E}}\mathcal{L}(\mathcal{E})
\approx 
\arg\min_{\mathcal{E}}
\mathbb{E}_{\substack{(X_i,Y_i)\sim\mathcal{D} \\ \varepsilon_i\sim\rho(\cdot\mid X_i,\mathcal{E})}}
\big[\mathcal{H}(Y_i \mid X_i, \varepsilon_i)\big].
\end{equation}
\end{corollary}
This perspective provides a more intuitive interpretation, \method{} learns to construct an experience library that reduces the model’s predictive uncertainty while aligning the model distribution with the underlying true distribution.

\textbf{Information-Theoretic Role of Experience.}
To further elucidate the role of retrieved experiences, we analyze their contribution from an information-theoretic perspective.
Specifically, the expected mutual information $\mathcal{I}$ between the target output $Y_i$ and the reasoning experience $\varepsilon_i\sim\rho(\cdot\mid X_i,\mathcal{E})$ is defined as:
\begin{equation}
\mathbb{E}_{\substack{(X_i,Y_i)\sim\mathcal{D} \\ \varepsilon_i\sim\rho(\cdot\mid X_i,\mathcal{E})}}
\big[\mathcal{I}(Y_i;\varepsilon_i\mid X_i)\big] =
\mathbb{E}_{\substack{(X_i,Y_i)\sim\mathcal{D} \\ \varepsilon_i\sim\rho(\cdot\mid X_i,\mathcal{E})}}
\big[\mathcal{H}(Y_i \mid X_i)\big] -
\mathbb{E}_{\substack{(X_i,Y_i)\sim\mathcal{D} \\ \varepsilon_i\sim\rho(\cdot\mid X_i,\mathcal{E})}}
\big[\mathcal{H}(Y_i \mid X_i, \varepsilon_i)\big].
\end{equation}
Since $\mathcal{H}(Y\mid X)$ remains constant for a given dataset, minimizing $\mathcal{H}(Y\mid X,\varepsilon)$ is equivalent to maximizing $\mathcal{I}(Y;\varepsilon\mid X)$. 
This implies that retrieved experiences $\varepsilon$ provide additional information about the target $Y$ beyond what is contained in the query $X$, thereby enhancing prediction confidence and shaping the conditional output distribution $\pi(Y\mid X,\varepsilon)$.

We can further quantify the contribution of informative experiences by comparing their information gain to irrelevant ones. 
Let $\varepsilon_i^+$ denote effective experiences and $\varepsilon_i^-$ denote irrelevant ones. 
The difference in conditional mutual information is:
\begin{equation}
\Delta \mathcal{I} = \mathcal{I}(Y_i;\varepsilon_i^+\mid X_i) - \mathcal{I}(Y_i;\varepsilon_i^-\mid X_i) = \mathcal{H}(Y_i\mid X_i,\varepsilon_i^-) - \mathcal{H}(Y_i\mid X_i,\varepsilon_i^+)
\end{equation}
The optimization objective in Corollary~\ref{cor:obj} guarantees $\mathcal{H}(Y_i\mid X_i,\varepsilon_i^+) < \mathcal{H}(Y_i\mid X_i,\varepsilon_i^-)$, thus $\Delta \mathcal{I} > 0$, indicating that effective experiences yield greater information gain.
To conclude, incorporating informative experiences substantially reduces predictive uncertainty, guiding the model toward more accurate and confident reasoning.


\begin{figure}[!t]
  \centering
  \includegraphics[width=0.98\linewidth]{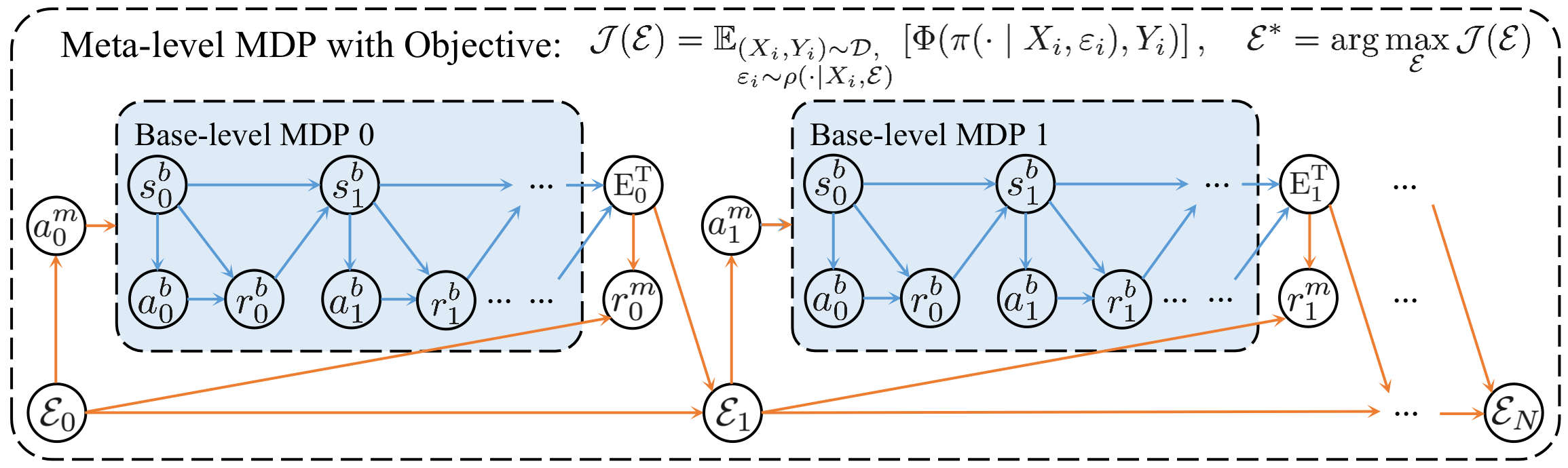}
  \caption{Illustration of the Meta-MDP formulation of \method{}.
The Base-level MDP performs intra-sample exploration and experience distillation, while the Meta-level MDP integrates these experiences to evolve the global experience library through forward updates.}
  \label{fig:mdp}
\end{figure}

\subsection{Optimize \method{} as Meta-MDP}
Unlike conventional gradient-based optimization, \method{} adopts a forward, experience-driven semantic evolution paradigm. 
Since the update rule of \method{} in Definition 2 produces an updated experience library that depends solely on the current state and is independent of historical states, we can formalize the optimization procedure as a hierarchical Meta Markov Decision Process (Meta-MDP) as shown in Fig~\ref{fig:mdp}, comprising Base-level MDPs that explores extensive experience, and a Meta-level MDP that governs the experience library evolution, enabling forwardly learning from accumulated experience instead of gradient back-propagation.

\paragraph{Meta-level MDP for Experience Library Evolution}
\begin{definition}[Meta-level MDP for Experience Library Evolution]
The Meta-level MDP is formally defined as the tuple $(\mathcal{S}^{m}, \mathcal{A}^{m}, \mathcal{T}^{m}, \mathcal{R}^{m})$:
\begin{itemize}
    \item \textbf{Meta-state} $s^{m}_i$: The current experience library $\mathcal{E}_i$
    \item \textbf{Meta-action} $a^{m}_i = \Psi({E^T_i}\mid X_i,Y_i,\mathcal{E}_i)$: Invokes the Base-level MDP on sample $(X_i,Y_i)$ under $\mathcal{E}_i$, yielding explored experiences ${E^T_i}$
    \item \textbf{Transition function} $\mathcal{T}^{m}$: Conducted by the updater $\mu$, evolves the experience library as $\mathcal{E}_{i+1}\sim\mathcal{T}^{m}(\mathcal{E}_{i+1}\mid \mathcal{E}_i,{E^T_i})$ under the meta-policy
    \item \textbf{Reward} $r^{m}_i = \Phi\big(\pi(\cdot\mid X_i, \rho(\cdot\mid X_i,\mathcal{E}_{i+1})),Y_i\big)$: Measures correctness of model output given $\mathcal{E}_{i+1}$
\end{itemize}
\end{definition}

Consequently, the cumulative return $G^{m} = \sum_{i=0}^{N-1} r^{m}_i$ over one meta-episode reflects the aggregated model accuracy across all $N$ training samples.
Therefore, maximizing $G^{m}$ is essentially equivalent to optimizing the objective in Eq.~\ref{eq:obj}.

In summary, this Meta-level formulation establishes a higher-order optimization loop that continuously refines the experience library through forward experiential updates, orchestrating Base-level exploration with long-term performance optimization.

\paragraph{Base-level MDP for Extensive Experiences Exploration} 
The Base-level MDP operationalizes the exploratory component of \method{} with an actor–critic framework. The actor iteratively generates diverse reasoning trajectories, while the critic provides semantic feedback that reflects on these trajectories, distilling them into structured experiences for Meta-level optimization, thus can be formulated as,
\begin{definition}[Base-level MDP for Experience Exploration]
The Base-level MDP is formally defined as the tuple $(\mathcal{S}^{b}, \mathcal{A}^{b}, \mathcal{T}^{b}, \mathcal{R}^{b})$:
\begin{itemize}
    \item \textbf{Base-state} $s^{b}_t$: The accumulated experience ${E^t_i}$ derived from the first $t$ reasoning trajectories explored on $X_i$ given $\mathcal{E}_i$
    \item \textbf{Base-action} $a^{b}_{t}$: One complete reasoning trajectory proposed by the actor $\pi$
    \item \textbf{Base-reward} $r^{b}_{t}=\Gamma(a^{b}_t,Y_i)$: Semantic feedback distilled by the critic, evaluating both the reasoning process and outcome relative to $Y_i$, capturing success patterns and interpretative insights
    \item \textbf{Transition function} $\mathcal{T}^{b}$: Updates the intra-sample experience accumulation as $s^{b}_{t+1} \sim \mathcal{T}^{b}(s^{b}_{t+1}\mid s^{b}_{t},r^{b}_{t})$, progressively integrating reflective knowledge until exploration terminates.
\end{itemize}
\end{definition}

Upon completion, the accumulated experience $E^T_i = s^{b}_T$ is returned to the Meta-level MDP for high-level experience library evolution.

Together, the Base-level and Meta-level MDPs form a hierarchical, closed-loop optimization process, the Base-level conducts fine-grained exploration and reflection within individual samples, while the Meta-level integrates these experiences across samples to continuously evolve the experience library in a gradient-free, forward-learning manner.

\section{Concrete Instantiation of \method{}}

\begin{figure}[!t]
  \centering
  \includegraphics[width=0.98\linewidth]{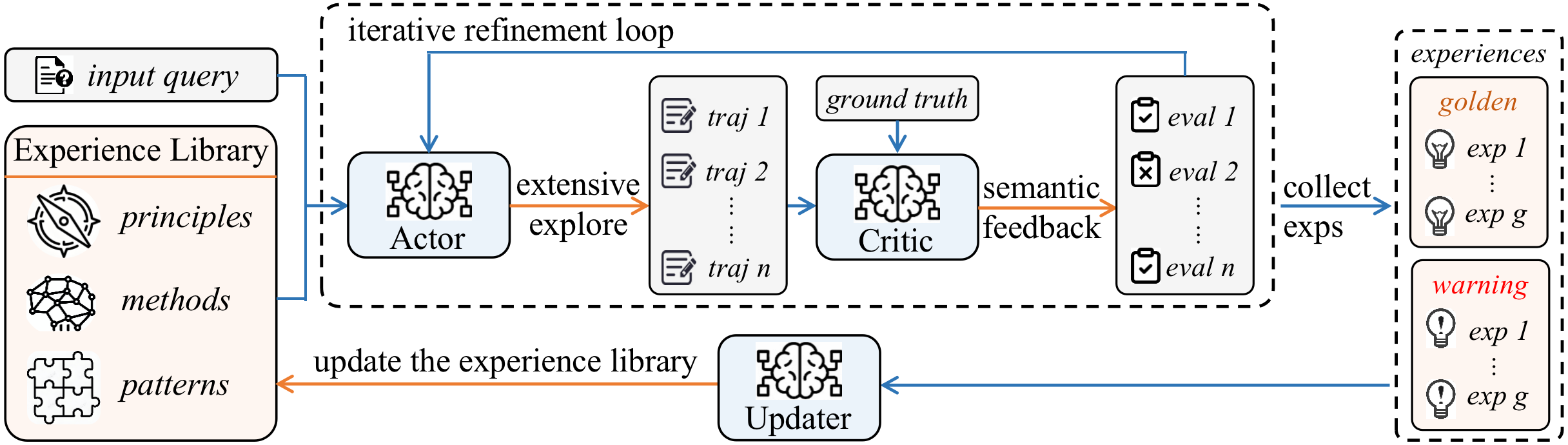}
  \caption{Concrete Instantiation of \method{}. The refinement loop of actor-critic iteratively explores and refines experiences, then the meta-level updater dynamically organizes the distilled experiences into the evolving experience library.}
  \label{fig:fl}
\end{figure}

Building on the mathematical formulation, we instantiate \method{} as a concrete mechanism consisting of two core components, namely, extensive experience exploration via the actor–critic forward loop and experience library evolution that consolidates and reuses experiential semantics.
Together, these two components realize continual semantic evolution through structured experience interaction as shown in Fig~\ref{fig:fl}.

\subsection{Extensive Experience Exploration}
The generation and collection of experiences are realized through per-case actor–critic interactions defined by the Base-level MDP. The key objective is to perform extensive exploration while ensuring that the collected experiences are both diverse and semantically informative.

\paragraph{Extensive Exploration Mechanism.}
To enable extensive high-quality exploration, we design a dual-scaling mechanism that combines parallel and sequential scaling. 
For \emph{parallel scaling}, multiple reasoning trajectories are sampled in parallel for the same input through rejection sampling~\cite{fujimoto2021minimalist,dong2023raft,gulcehre2023reinforced}, allowing diverse reasoning hypotheses to be explored and high-quality outputs to be retained.
For \emph{sequential scaling}, the actor iteratively improves upon its previous output under the critic’s semantic feedback~\cite{yuksekgonul2024textgrad,shinn2023reflexion}, where the critic provides constructive guidance and refinement cues derived from the prior round.
This cooperative exploration strategy balances breadth and depth, ensuring both coverage of alternative reasoning paths and progressive improvement over iterations.

\paragraph{Iterative Refinement with Semantic Signal.}
To ensure that collected experiences contribute meaningfully to learning, we design an iterative refinement mechanism driven by semantic evaluation.
For reasoning trajectories that lead to incorrect conclusions, the critic first identifies the underlying causes of failure and summarizes them into abstract, task-agnostic improvement suggestions.
These generalized feedback signals, denoted as new experiences $E_t$, are fed back into the actor together with its previous reasoning trace to initiate the next actor–critic iteration.
This process continues until the actor produces a correct or sufficiently improved reasoning trajectory, or a predefined iteration limit is reached.
In this loop, the critic’s feedback functions as a \emph{semantic update signal}~\cite{yuksekgonul2024textgrad,shinn2023reflexion,madaan2023self} that guides the actor’s reasoning refinement, analogous to the gradient signal in traditional optimization, yet expressed in natural language semantics rather than numerical differentials.
Unlike backpropagation-based learning, which relies on scalar gradients with limited interpretive bandwidth and heavy computational budget, this forward semantic feedback mechanism provides rich interpretable guidance without costly parameter updating, aligning with the cognitive characteristics of \textit{human-like learning from experience}.

Once the iterative refinement completes, validated experiences are distilled and passed to the Meta-level MDP, evolving experience library for long-term accumulation.

\subsection{Experience Library Evolution}
The experience library is hierarchically organized and continually evolves within the Meta-level MDP formulated above, driven by two complementary operations, update and retrieval, which together constitute the core meta-level transition dynamics.

\paragraph{Hierarchical Experience Library.}
The experience library adopts an explicit textual storage mechanism, which enables explicit recording of experiential rules, externalization of internal reasoning behaviors, and visualization of the learning dynamics, thereby enhancing interpretability and transparency.
To effectively organize learned experiences varying in granularity, abstraction, and contextual dependency, we design a hierarchical organization mechanism with three interconnected levels, namely, high-level strategic principles and guidelines, mid-level reasoning patterns and methodological templates, and low-level factual knowledge and concrete instances.
In addition, the library is partitioned into two complementary zones: the \textit{golden} zone, which stores experiences distilled from correct reasoning trajectories, and the \textit{warning} zone, which records failure cases and diagnostic insights extracted from erroneous trajectories.
Such dual-zone organization facilitates balanced learning from both successes and failures, allowing the agent to not only consolidate validated reasoning paths but also internalize the lessons encoded in previous mistakes.
This structure allows experiences to be modularly indexed and retrieved across abstraction levels, enabling both top-down guidance and bottom-up consolidation of knowledge.
Hierarchical organization also supports dynamic restructuring as new experiences are incorporated, maintaining the coherence of the library during continual evolution.

\paragraph{Experience Update Mechanism.}
Given a newly generated experience $\varepsilon$, the \textbf{updater} agent autonomously decides whether and how to incorporate it into the experience library. 
If an identical entry already exists, $\varepsilon$ is discarded to maintain compactness. 
If semantically similar entries are found, the agent performs a selective merge, preserving the more informative or higher-quality trajectory while minimizing redundancy. 
Otherwise, $\varepsilon$ is inserted into the appropriate partition and hierarchical level according to its semantic granularity. 
This adaptive update process ensures that the experience library incrementally evolves toward a more structured and informative representation of accumulated knowledge, rather than expanding indiscriminately or degenerating into redundant memory accumulation.

\paragraph{Experience Retrieval Mechanism.}
During inference, the experience library is encapsulated as an interactive \textbf{tool} callable by the agent~\cite{yao2022react,qiu2025alita}.
Instead of static vector-based semantic-similarity search~\cite{reimers2019sentence,ram2023context,karpukhin2020dense}, the agent performs contextualized retrieval conditioned on the specific query and its current reasoning state. 
This allows the agent to interpret relevance not only by semantic similarity but also by contextual consistency and goal alignment, thereby avoiding the pitfalls of retrieving semantically similar yet pragmatically contradictory entries.
Retrieval proceeds hierarchically, the agent first identifies relevant high-level strategies, then locates corresponding procedural patterns, and finally accesses concrete case examples within each pattern. 
At each stage, the top-$k$ (typically $k=5$) most relevant entries are selected to balance precision and efficiency. 
The retrieved experiences support the agent’s ongoing reasoning, reflective self-evaluation, and corrective refinement.
Moreover, retrieval is not restricted to pre-inference access, instead, the agent can dynamically query the library during reasoning, enabling adaptive integration of prior knowledge into evolving cognitive trajectories.

Overall, the update–retrieval dynamics establish a self-organizing experiential system that progressively refines its hierarchical organization and semantic granularity, facilitating the agent’s continual evolution through accumulated experience.

    \section{Experiments}
We evaluate the \method{} paradigm across a diverse suite of challenging environments, testing its effectiveness, and demonstrating the scaling law, transferability of the learned experience library, and a thorough case study in all settings.

\subsection{Experimental Setup}
Our experimental setup are structured to comprehensively evaluate the performance and generalizability of \method{}, involving diverse benchmarks, comparisons against established baselines, and a detailed illustration of the training procedures.

\paragraph{Evaluation Benchmarks.}
We evaluate the proposed \method{} paradigm across four challenging benchmarks spanning three scientific domains (\textit{i.e.}, mathematics, chemistry, and biology) to assess both reasoning capability and domain adaptability. Specifically: (1) AIME25 competition dataset~\cite{aime2025problems} for complicated mathematical reasoning; (2) GSM8k dataset~\cite{cobbe2021gsm8k} multi-step arithmetic reasoning; (3) USPTO50k~\cite{schneider2016s} benchmark for canonical single-step retrosynthesis; (4) ProteinGym~\cite{notin2023proteingym} benchmark for protein fitness prediction. 

\paragraph{Compared Methods.}
Our evaluation spans a range of base models, tested on four distinct experimental configurations for comparison: (1) Vanilla LLM, (2) LLM with In-Context Learning (ICL)~\cite{dong2022survey}, (3) LLM agent with reasoning-and-acting workflow (Agent)~\cite{yao2023reactsynergizingreasoningacting}, and (4) our proposed \method{}. For \method{}, each model constructs an experience library during training, which is later utilized at test-time. The library structure is adapted to the benchmark: hierarchical  for AIME25 and USPTO50k, and non-hierarchical for GSM8k and ProteinGym.

\paragraph{Training Configurations.}
For AIME25, the training was conducted on 49 historical problems from AIME83 to AIME24. On GSM8k, we adhere to the official training and testing splits~\cite{cobbe2021gsm8k}. For USPTO50k, 100 instances are randomly sampled from the original 5k test split for evaluation, and 50 instances from the training split are randomly sampled for training. For ProteinGym, 100 mutational sequences are sampled for each wild-type protein target as training set, which is on average 1.47\% of the available data. This design reflects practical constraints in protein engineering, where only a limited number of labeled mutations are known, while the vast majority of sequence space remains unexplored. 

\subsection{Main Results}
Our main results demonstrate that \method{} achieves significant and consistent improvements across all scientific domains and benchmarks, as detailed below.


\begin{table}[!t]
\centering
\caption{Results on four benchmarks. All values are accuracies in percentage, which is accuracy for AIME25, GSM8k, USPTO50k and spearman's $\rho$ for ProteinGym. ICL stands for the performance of LLM with in context learning, and Agent indicates the performance of LLM agent with ReAct workflow.}
\label{tab:main_table}
\resizebox{\textwidth}{!}{%
\begin{tabular}{llcccc}
\toprule
\textbf{Benchmark} & \textbf{Models} & \textbf{LLM} & \textbf{ICL} & \textbf{Agent} & \textbf{\method{}} \\
\midrule
\multicolumn{6}{c}{\textit{\textbf{Math}}} \\
\midrule
\midrule
\multirow{2}{*}{\textbf{AIME25}} & \textit{Claude-Sonnet-4}~\cite{anthropic2025claude4} & 40.0 & 30.0 (\textcolor{red!80}{-10.0}) & 50.0 (\textcolor{blue!80}{+10.0}) & \textbf{63.3 (\textcolor{blue!80}{+23.3})} \\
& \textit{DeepSeek-V3.1-Terminus}~\cite{DeepSeekV3.1Terminus} & 56.7 & 53.3 (\textcolor{red!80}{-3.3}) & 60.0 (\textcolor{blue!80}{+3.3}) & \textbf{66.6 (\textcolor{blue!80}{+10.0})} \\
\midrule
\multirow{4}{*}{\textbf{GSM8k}} & \textit{GPT-3.5}~\cite{OpenAI_GPT3.5_Turbo_docs} & 80.8 & 81.4 (\textcolor{blue!80}{+0.6}) & 78.5 (\textcolor{red!80}{-2.3}) & \textbf{83.3 (\textcolor{blue!80}{+3.3})} \\
& \textit{GPT-4}~\cite{openai2024gpt4technicalreport} & 93.8 & 94.2 (\textcolor{blue!80}{+0.4}) & 94.0 (\textcolor{blue!80}{+0.2}) & \textbf{95.9 (\textcolor{blue!80}{+2.1})}  \\
& \textit{Llama-3.2-1B}~\cite{MetaLlama3.2_2024} & 74.3 & 75.8 (\textcolor{blue!80}{+1.5}) & 71.0 (\textcolor{red!80}{-3.3}) & \textbf{80.9 (\textcolor{blue!80}{+6.6})}  \\
& \textit{Llama-3.2-3B}~\cite{MetaLlama3.2_2024} & 78.4 & 78.8 (\textcolor{blue!80}{+0.4}) & 74.5 (\textcolor{red!80}{-3.9}) & \textbf{81.1 (\textcolor{blue!80}{+2.7})}  \\
\midrule
\multicolumn{6}{c}{\textit{\textbf{Chemistry}}} \\
\midrule
\midrule
\multirow{3}{*}{\textbf{USPTO50k}} & \textit{GPT-5}~\cite{openai2025gpt5} & 9.0 & 14.0 (\textcolor{blue!80}{+5.0}) & 12.0 (\textcolor{blue!80}{+3.0}) & \textbf{16.0 (\textcolor{blue!80}{+7.0})} \\
& \textit{Gemini-2.5-Pro}~\cite{google2025gemini2_5_pro} & 9.0 & 15.0 (\textcolor{blue!80}{+6.0}) & 12.0 (\textcolor{blue!80}{+3.0}) & \textbf{18.0 (\textcolor{blue!80}{+9.0})}  \\
& \textit{Claude-Sonnet-4.5}~\cite{anthropic2025sonnet4.5} & 20.0 & 24.0 (\textcolor{blue!80}{+4.0}) & 23.0 (\textcolor{blue!80}{+3.0}) & \textbf{30.0 (\textcolor{blue!80}{+10.0})}  \\
\midrule
\multicolumn{6}{c}{\textit{\textbf{Biology}}} \\
\midrule
\midrule
\multirow{3}{*}{\textbf{ProteinGym}} & \textit{DeepSeek-V3.1-Terminus}~\cite{DeepSeekV3.1Terminus} & 47.9 & 48.9 (\textcolor{blue!80}{+1.0}) & 48.6 (\textcolor{blue!80}{+0.7}) & \textbf{56.8 (\textcolor{blue!80}{+8.9})} \\
& \textit{GPT-OSS-120B}~\cite{openai2025gptoss120bgptoss20bmodel} & 47.7 & 49.8 (\textcolor{blue!80}{+2.1}) & 51.5 (\textcolor{blue!80}{+3.8}) & \textbf{57.3 (\textcolor{blue!80}{+9.6})}  \\
& \textit{Claude-Sonnet-4}~\cite{anthropic2025claude4} & 46.0 & 49.8 (\textcolor{blue!80}{+3.8}) & 50.2 (\textcolor{blue!80}{+4.2}) & \textbf{59.7 (\textcolor{blue!80}{+13.7})}  \\
\bottomrule
\end{tabular}
}
\end{table}


\paragraph{Math.} As shown in Table~\ref{tab:main_table} and Figure~\ref{fig:main_method_overview}, our \method{} paradigm consistently and significantly outperforms all baselines across both mathematical reasoning benchmarks. On the challenging AIME25 dataset, \method{} boosts the accuracy of Claude-Sonnet-4 from a $40.0\%$ baseline to an impressive $63.3\%$ after learning from only 49 examples. This substantial $58.7\%$ relative improvement demonstrates its powerful learning capability. Similarly, on GSM8k, \method{} consistently enhances the performance of already strong models, such as elevating GPT-4's accuracy to $95.9\%$. These results demonstrate that our parameter-free paradigm can yield substantial gains in reasoning, an advancement previously achieved primarily through parameter-based learning methods~\cite{deepseekai2025deepseekr1incentivizingreasoningcapability,yu2025dapoopensourcellmreinforcement}.

\paragraph{Chemistry.} In the specialized domain of chemistry, \method{} proves to be highly effective at bridging the performance gap of generalist models. On the USPTO50k benchmark, baseline models exhibit limited proficiency, with GPT-5 and Gemini-2.5-Pro achieving only $9.0\%$ accuracy. By learning from a small set of 50 examples, \method{} nearly doubles these scores, increasing them to $16.0\% (+7.0\%)$ and $18.0\% (+9.0\%)$, respectively. The most notable gain is observed with Claude-Sonnet-4.5, which improves from $20.0\%$ to $30.0\%$ accuracy (+$10.0\%$). These results underscore the efficacy of \method{} as a sample-efficient, parameter-free approach for adapting models to complex scientific domains.

\begin{figure}[!t]
  \centering
  \includegraphics[width=0.98\linewidth]{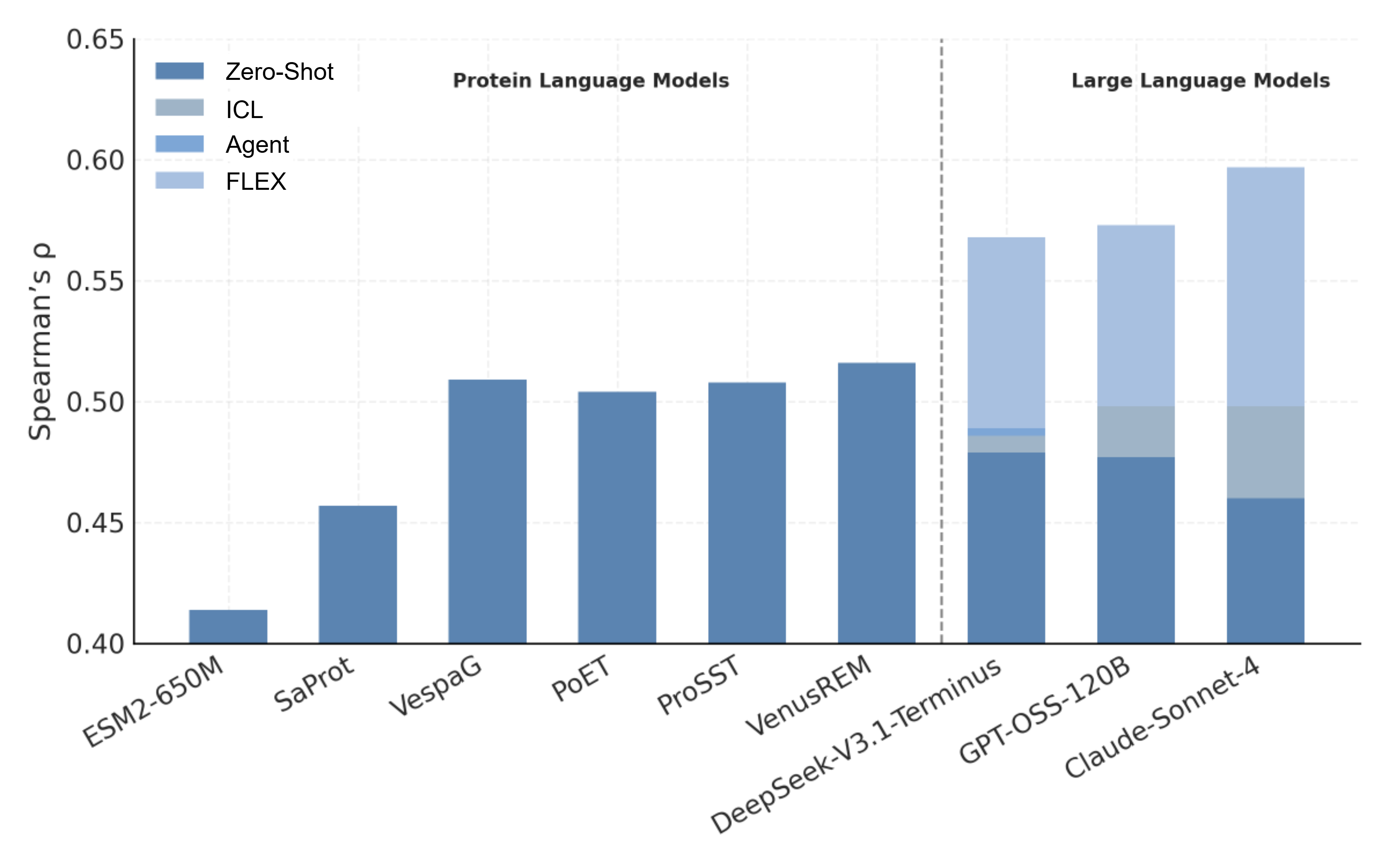}
  \caption{ProteinGym Results from both protein language model and large language model.}
  \label{fig:proteingym_all}
\end{figure}

\paragraph{Biology.} For the ProteinGym zero-shot benchmark, the prevailing state-of-the-art Spearman’s correlation remains near 0.52. As shown in Table~\ref{tab:main_table} and Figure~\ref{fig:proteingym_all}, off-the-shelf large language models (LLMs) tend to underperform relative to specialized protein language models even conducting subtask of identifying and appropriately combining functionally important positions. This shortfall persists even when LLMs are supplemented with simple in-context examples, which suggests that mere exposure to a few labeled instances is insufficient to endow these models with the inductive biases required for robust mutational effect prediction. Importantly, we find that a forward-learning paradigm substantially mitigates this limitation. By extracting compact, generalizable “experience” from a small training set and encoding it as reusable rules or heuristics, the forward-learned model is able to transfer these insights to a much larger and more diverse test set — yielding an average improvement of roughly 0.10 in Spearman’s $\rho$.


\subsection{The Scaling Law of Experience}

\begin{figure}[!t]
  \centering
  \includegraphics[width=0.98\linewidth]{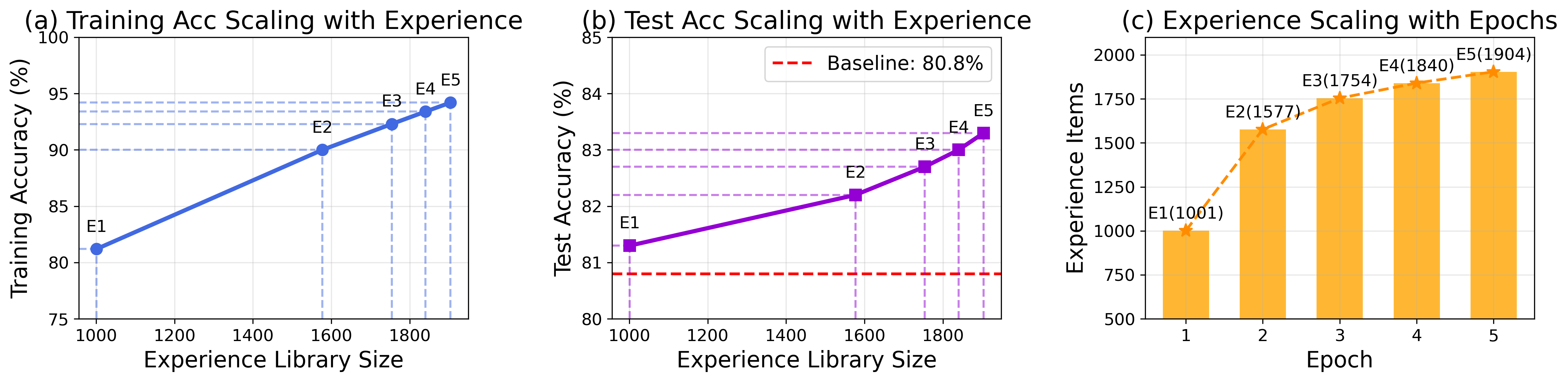}
  \caption{Training dynamics and scaling laws of \method{} on the GSM8K dataset across 5 epochs. Training accuracy and test accuracy both show strong scalability with the size of the experience library. Experience library also exhibits scaling law with the epochs.}
  \label{fig:td}
\end{figure}

To understand how experience accumulation influences model performance, we systematically analyze the scaling properties of \method{}. Our investigation reveals a predictable and principled relationship between the size of the experience library and model capability. As illustrated in Fig.~\ref{fig:td}, we identify three distinct scaling regimes through rigorous analysis across 5 training epochs on GSM8k.

\textbf{Training Accuracy Scaling with Experience.} Figure~\ref{fig:td}(a) shows a strong power-law relationship, with training accuracy increasing from 81.2\% to 94.2\% as the experience library grows from 1,001 to 1,904 items. This demonstrates substantial returns from experience accumulation, with performance gains following a predictable scaling trajectory.

\textbf{Generalization Scaling with Experience.} The test accuracy in Figure~\ref{fig:td}(b) exhibits consistent improvement from 81.3\% to 83.3\%, significantly surpassing the 80.8\% baseline. The reduced variance with increasing experience indicates that accumulated knowledge stabilizes predictions and enhances reasoning robustness.

\textbf{Experience Scaling with Epochs.} Figure~\ref{fig:td}(c) uncovers a distinct scaling regime for experience accumulation itself. The process follows a logistic-like curve: rapid initial expansion (+576 experiences from epoch 1 to 2) transitions to selective refinement (+64 from epoch 4 to 5). This indicates an intelligent, phase-changing scaling law that efficiently covers the problem space while strategically avoiding redundancy.

These scaling laws establish \method{} as a principled framework for continuous model improvement, offering predictable performance gains through experience accumulation. The consistent scaling relationships across training and generalization metrics provide strong empirical grounding for experience-driven learning paradigms, suggesting new pathways for model enhancement that transcend traditional scaling approaches based solely on parameter count or compute budget to the \textbf{scaling with experience}.

\subsection{Inheritance of the Experience Library}
\begin{table}[!t]
\centering
\caption{Performance comparison on cross-model memory transfer across two benchmarks (AIME25 and USPTO50k). Values in parentheses denote the absolute accuracy improvement over the ReAct baseline for each model. The models in columns represent the source of inherited experience. For AIME25, Claude and DeepSeek in columns stand for Claude-Sonnet-4 and DeepSeek-V3.1-Teminus. For USPTO50k, GPT, Gemini, and Claude in columns stand for GPT-5, Gemini-2.5-Pro, and Claude-Sonnet-4.5. For ProteinGym, Qwen and DeepSeek in columns stand for Qwen3-8B and DeepSeek-V3.1-Terminus.}
\label{tab:memory_transfer}
\smallskip
\begin{tabular}{@{}lcccc@{}}
\toprule
\multicolumn{5}{c}{\textbf{AIME25}} \\
\midrule
\textbf{Model} & \textbf{Agent} & \multicolumn{2}{c}{\textbf{+ Claude}} & \textbf{+ DeepSeek} \\
\midrule
\textit{Claude-Sonnet-4} & 50.0 & \multicolumn{2}{c}{63.3 (\textcolor{blue!80}{+13.3})}  & 66.7 (\textcolor{blue!80}{+16.7}) \\
\textit{DeepSeek-V3.1-Terminus}  & 60.0  & \multicolumn{2}{c}{66.7 (\textcolor{blue!80}{+6.7})}  & 66.7 (\textcolor{blue!80}{+6.7}) \\
\midrule
\multicolumn{5}{c}{\textbf{USPTO50k}} \\
\midrule
 {\textbf{Model}} &  \textbf{Agent} &  {\textbf{+ GPT}} & {\textbf{+ Gemini}} &  {\textbf{+ Claude}} \\
\midrule
 {\textit{GPT-5}} &  {12.0} &  {16.0 (\textcolor{blue!80}{+4.0)} } &  18.0 (\textcolor{blue!80}{+6.0}) &  {15.0 (\textcolor{blue!80}{+3.0)} }\\
 {\textit{Gemini-2.5-Pro} } &  {12.0} &  {14.0 (\textcolor{blue!80}{+2.0)} } &  18.0 (\textcolor{blue!80}{+6.0})&  {23.0 (\textcolor{blue!80}{+11.0)} }\\
 {\textit{Claude-Sonnet-4.5} } &  {23.0} &  {28.0 (\textcolor{blue!80}{+5.0)} } &  30.0 (\textcolor{blue!80}{+7.0}) &  30.0 (\textcolor{blue!80}{+7.0)} \\
\midrule
\multicolumn{5}{c}{\textbf{ProteinGym}} \\
\midrule
\textbf{Model} & \textbf{Agent}& \multicolumn{2}{c}{\textbf{+ Qwen}} & \textbf{+ DeepSeek\quad\  }\\
\midrule
\textit{GPT-OSS} & {51.5} & \multicolumn{2}{c}{55.0 (\textcolor{blue!80}{+3.5})}  & 56.6 (\textcolor{blue!80}{+5.1})\\
\toprule
\end{tabular}%
\end{table}

A key advantage of \method{} over parameter-based approaches is its exceptional inheritance property (Fig~\ref{fig:inheritance}). By decoupling learned knowledge from model weights, the experience library acts as a lightweight, plug-and-play module that can be seamlessly transferred across different agents. Our experiments in Table~\ref{tab:memory_transfer} reveal two significant and economically valuable properties: the \textbf{distillation} of expertise from strong to weaker models, and the \textbf{generality} of strategies from weaker to stronger models.

First, we observe a powerful distillation effect, where expertise from a top-performing model can significantly uplift weaker ones. For instance, on USPTO50k, the experience library generated by the strongest model, Claude-Sonnet-4.5, boosts the performance of Gemini-2.5-Pro by an impressive 11 absolute points. This highlights a cost-effective pathway: the expensive exploration of a single, powerful agent can be captured and reused to enhance a fleet of less capable models without the need for individual fine-tuning. A similar trend is evident on AIME25, where the experience from the more capable DeepSeek-V3.1-Terminus model provides the largest performance gain (+16.7\%) for Claude-Sonnet-4.

Even more strikingly, our results demonstrate a strong generality effect, where experience from weaker models provides substantial benefits to stronger ones. Notably, on AIME25, the experience library from the weaker Claude-Sonnet-4 elevates the stronger DeepSeek-V3.1-Terminus by 6.7 points, achieving the exact same performance level as when DeepSeek uses its own meticulously generated experience. This finding is profound: it proves that the learned experience captures fundamental, high-level strategies that are model-agnostic, rather than model-specific artifacts. The same trend is corroborated on USPTO50k and ProteinGym, where experiences from weaker models(Qwen3-8B, DeepSeek-V3.1-Terminus, GPT-5, Gemini-2.5-Pro) substantially improve the already capable Claude-Sonnet-4.5 or GPT-OSS-120B.

Taken together, these phenomena establish the experience library as a truly portable knowledge module. This opens a path towards a new paradigm: creating a single, universal experience library through a one-time training process, or even mixing experiences from diverse sources, to provide general performance improvements across a wide ecosystem of agents.

\begin{figure}[!htbp]
    \centering
    \includegraphics[width=0.8\linewidth]{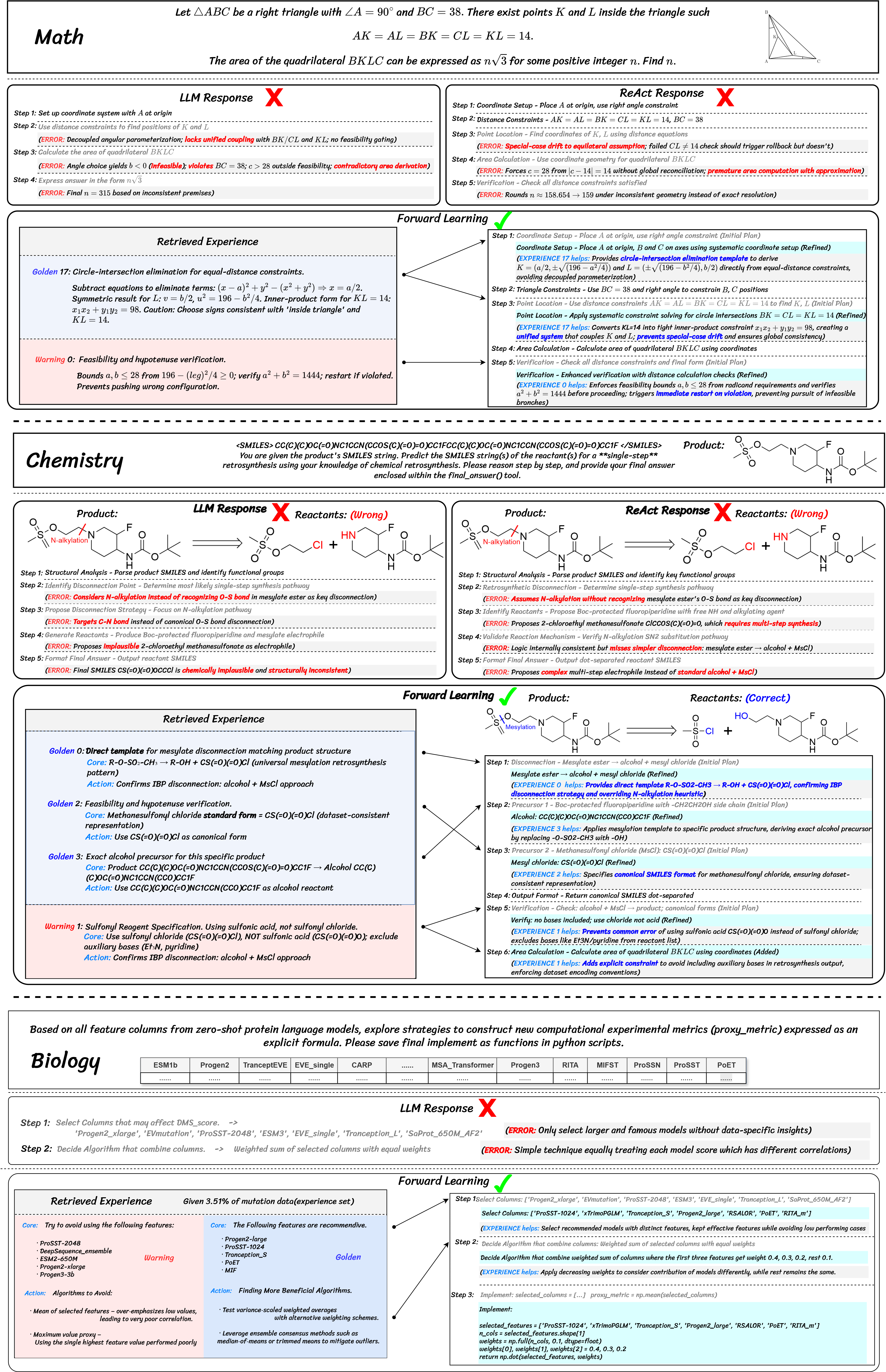}
    \caption{Qualitative case studies in Mathematics, Chemistry, and Biology demonstrating the effectiveness of \method{}. In each domain, baseline agents (LLM Response, ReAct Response) fail due to critical reasoning errors (marked with \textcolor{red}{\ding{55}}). In contrast, by retrieving and applying distilled knowledge (\textit{e.g.}, Golden rules and Warnings) from its experience library, \method{} successfully refines its strategy, overcomes the initial failures, and arrives at the correct solution (marked with \textcolor{green!70!black}{\ding{51}}).}
    \label{fig:case_study}
\end{figure}

\subsection{Case Study}
This section presents case studies across three domains to illustrate how the learned experience library by \method{} corrects initial failures by providing critical procedural knowledge and transforms flawed reasoning into correct solutions.

\paragraph{Math.} The distinct problem-solving trajectories in Figure~\ref{fig:case_study} highlight the transformative impact of our \method{} paradigm. The naive LLM fails by decoupling geometric constraints, leading to infeasible side lengths. The standard ReAct agent, while more structured, drifts into special-case assumptions without verifying global consistency. In stark contrast, \method{} leverages its experience library to inject structured, procedural knowledge into the reasoning cycle. By retrieving a reusable algebraic template and critical feasibility checks (e.g., $a^2 + b^2 = 1444$, $a,b \le 28$), it directly addresses the core LLM limitations of logical drift and constraint violation. This distilled experience transforms the agent's process from a heuristic-driven exploration into a deterministic, solve-verify loop, pruning invalid branches and enforcing global reconciliation. Essentially, the experience library provides the procedural scaffolds and logical guardrails that the other approaches lack, enabling the agent to systematically find the correct parameters and compute the final area of $104\sqrt{3}$. This demonstrates how our paradigm bridges the gap between an LLM's declarative knowledge and the effective procedural execution required for complex problem-solving.

\paragraph{Chemistry.} The trajectories in Figure~\ref{fig:case_study} reveal a critical gap in the domain-specific reasoning of standard models. Both the naive LLM and the vanilla ReAct agent make the same fundamental error: despite correctly identifying the mesylate group, they misidentify the disconnection point by prioritizing a plausible but incorrect N-alkylation pathway. This demonstrates a failure to translate declarative knowledge (what a mesylate is) into correct procedural execution (how to disconnect it). In stark contrast, \method{} succeeds by leveraging its experience library to bridge this gap. It retrieves an explicit template ($\mathrm{R{-}O{-}SO_2{-}CH_3} \rightarrow \mathrm{R{-}OH} + \mathrm{CS(=O)(=O)Cl}$) that overrides the flawed heuristic and enforces the canonical $\mathrm{O{-}S}$ bond disconnection. The experience library thus functions as an external knowledge substrate, injecting proven procedural scaffolds and domain-specific constraints into the agent's workflow. This transforms the reasoning process from speculative hypothesis testing into a deterministic, domain-grounded execution, ensuring the correct retrosynthesis is identified.

\paragraph{Biology.} Figure~\ref{fig:case_study} provides an example of how agents learn from a deliberately designed experience library—golden rules, warnings, and core know-how. Unlike the math and retro-synthesis problems above, protein-fitness prediction rarely admits absolute right/wrong answers; instead, useful guidance must be distilled from empirical trial trajectories and cross validation signals. The library captures those trajectories as reusable micro-strategies and failure patterns, enabling the agent to adapt a generic regression objective to the idiosyncrasies of a specific protein. In practice this means shifting attention away from globally renowned, aggregate metrics toward complementary, target-specific features and modeling choices that actually improve performance on the individual proteins. Moreover, the distilled tips actively drive algorithmic evolution: they suggest pre-processing pipelines, more detailed feature-combination heuristics, and validation gates that make final formula more robust. By overlaying proven procedural scaffolds onto the frozen LLM’s answer, the experience library encourages more valuable and diverse explorations and simultaneously enlarges meaningful solutions that better tuned to the peculiarities scenarios.

    \section{Discussion}
\begin{figure}[!t]
    \centering
    \includegraphics[width=\linewidth]{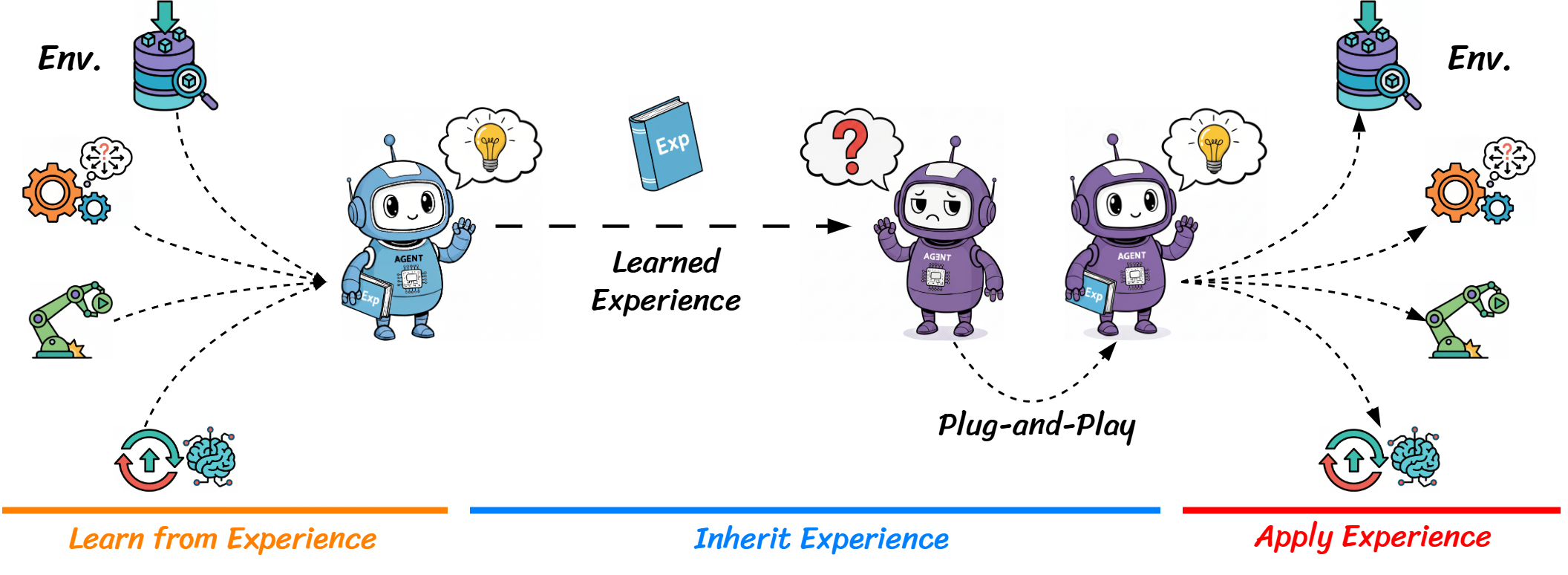}
    \caption{The process of inheriting the experience library across diverse agents. The experience library acts as a plug-and-play module that can boost agents' performance with a single training procedure.}
    \label{fig:inheritance}
\end{figure}

The aspiration to create intelligent agents that evolve through lived experience is a long-standing and central theme in artificial intelligence~\cite{turing2009,Russell2021AIMA}. This discussion aims to connect our research to this enduring quest, reflecting on the potential futures it might suggest and the historical context from which it emerges. We hope to offer a thoughtful perspective on the path ahead, framing our contributions not as conclusions, but as a humble step in an ongoing, collective journey.

\paragraph{Toward Continuous Learning.}
A central goal in AI is to enable agents to learn continuously from their interactions, much like living organisms do~\cite{sutton2023albertaplanairesearch,wang2024comprehensivesurveycontinuallearning}. This pursuit hints at a future where AI systems are not static artifacts, but dynamic partners capable of adapting in real time. For years, the research community has explored various avenues toward this goal, often seeking learning paradigms that are more "forward-looking" than traditional back-propagation~\cite{hinton2022forwardforwardalgorithmpreliminaryinvestigations}. The semantic capabilities of modern LLMs~\cite{openai2025gpt5,anthropic2025sonnet4.5,anthropic2025claude4,xai2025grok4}, however, offer a new lens for this quest. It becomes possible to explore learning not only as a numerical optimization task but as a process of reasoned reflection. Our explorations in this work suggest this is a worthwhile direction, raising the possibility for a new generation of AI systems whose learning processes are inherently more auditable and aligned with a lifelong learning model.

\paragraph{Heading to a Collective Wisdom.}
Beyond individual capability, another grand challenge is fostering a form of collective intelligence, mirroring how human progress is built upon shared knowledge~\cite{10.1145/3368986}. One can envision a future where a global network of specialized agents collaborate, forming a collective scientific mind to accelerate progress on humanity's greatest challenges. A key historical obstacle to this vision has been the absence of a robust medium to inherit wisdom across distinct AI agents~\cite{CAMPBELL200257,silver2017masteringchessshogiselfplay}. The concept of decoupling learned experience from an agent's internal parameters represents a hopeful path forward. Our explorations in this direction (Fig~\ref{fig:inheritance}) suggest that distilled experience can indeed serve as a viable medium for inheritance, adding a small contribution to the larger effort of building a foundation upon which a more synergistic AI ecosystem might one day be built.

\paragraph{Learning in a transparent way.}
Integral to the future of AI is the development of systems whose reasoning is transparent. This points toward a future where human-AI collaboration is built on a foundation of shared understanding, which is crucial for their integration into critical societal functions. The "black box" nature of many deep learning models has presented a persistent barrier to this goal~\cite{https://doi.org/10.1002/ail2.61,gohel2021explainableaicurrentstatus,hsieh2024comprehensiveguideexplainableai}. An experience-driven learning paradigm inherently addresses this challenge. When an agent's growth is chronicled as a series of explicit, human-readable experiences, its evolution becomes an open book. This transparency provides a direct mechanism for meaningful human-in-the-loop interaction~\cite{Wu_2022}, allowing experts to understand, guide, and even enrich an agent's learning journey. It represents a step toward a more white-box model of AI development, where the process of learning is as important as the final performance.

In conclusion, the ideas explored here are offered as a contribution to an ongoing conversation. It is our sincere hope that this work serves as one humble effort for the field, contributing to a future where AI evolves as a more dynamic, collaborative, and transparent partner in the human endeavor.

    \section{Related Work}
\subsection{Learning Paradigms}
Supervised fine-tuning (SFT) adapts LLMs to downstream distributions via gradient-based optimization on labeled data or human-written instuctions~\cite{howard2018universal,devlin2019bert,radford2019language,raffel2020exploring}.
It originates from the pretrain-finetune paradigm established by ULMFiT~\cite{howard2018universal}, BERT~\cite{devlin2019bert} and GPT-2~\cite{radford2019language}, and was later formalized as the first stage of alignment training in InstructGPT~\cite{ouyang2022training} and FLAN~\cite{wei2021finetuned}.
SFT further demonstrates remarkable progress across multiple fronts, including instruction following~\cite{ouyang2022training,sanh2021multitask,wei2021finetuned}, problem solving~\cite{toshniwal2024openmathinstruct,ahn2024large,luo2023wizardmath}, domain-specific adaptation~\cite{luo2022biogpt,zhang2024chemllm,liu2023fingpt,zhou2024lawgpt}, and so on~\cite{chen2021evaluating,roziere2023code,li2023starcoder,fried2022incoder}.

Reinforcement learning (RL) further optimizes LLMs through policy gradient~\cite{schulman2017proximal,sutton1999policy,schulman2015trust}, aligning model behaviors with human preferences or task-specific rewards, as exemplified by RLHF~\cite{christiano2017deep,ziegler2019fine,stiennon2020learning,ouyang2022training,rafailov2023direct} and RLAIF~\cite{bai2022training,lee2023rlaif}
Beyond alignment, RL has recently proven effective in enhancing model reasoning capabilities on challenging tasks, such as mathematical reasoning~\cite{shao2024deepseekmath,guo2025deepseek,yu2025dapo,yue2025vapo}, code generation~\cite{le2022coderl,shojaee2023execution,wang2024rlcoder}, as well as broader structured decision-making tasks~\cite{cui2025entropy,hu2025open,zheng2025group,zhang2025survey}.

Despite the effectiveness, gradient-based learning paradigms suffer from heavy computational cost and catastrophic forgetting~\cite{goodfellow2013empirical,hadi2023large}, and the resulting models remain static after training, unable to continually evolve through interaction with the environment~\cite{abbas2023loss}.

Non-parametric adaptation, including prompt engineering~\cite{sahoo2024systematic} and in-context learning~\cite{dong2022survey}, is another paradigm without gradient-based optimization.
Prompt engineering focuses on explicitly designing task instructions or linguistic cues that steer model outputs through semantic conditioning, eliciting step-by-step reasoning without parameter updates~\cite{kojima2022large,reynolds2021prompt,wang2022self,yao2023tree,besta2024graph,sahoo2024systematic}.
In-context learning adapts frozen LLMs by conditioning them on exemplars of input–output pairs or intermediate reasoning trajectories~\cite{brown2020language,wei2022chain,dong2022survey}.
However, these works focus on optimizing prompt or context, failing to make model learn from the experiences through interaction with environment.

\subsection{Self-Evolving Agent}
Recent research has begun to explore the vision of self-evolving agents~\cite{gao2025survey,fang2025comprehensive}, that can autonomously improve their reasoning and decision-making through continual interaction with the environment.
A first line of work focuses on tool evolution, where agents autonomously master, create, and utilize tools to extend their capabilities, such as Toolformer~\cite{schick2023toolformer}, ReAct~\cite{yao2022react}, Voyager~\cite{wang2023voyager}, CREATOR~\cite{qian2023creator}, and ALITA~\cite{qiu2025alita}.
Another line emphasizes agents architecture evolution, such as CAMEL~\cite{li2023camel}, MetaGPT~\cite{hong2023metagpt}, AgentSquare~\cite{shang2024agentsquare}, MaAS~\cite{zhang2025multi}, AutoFlow~\cite{li2024autoflow}, and GPTSwarm~\cite{zhuge2024gptswarm}, which orchestra multiple agents for cooperative problem solving.
A third line focuses on context evolution, where agents continually refine their internal reasoning via reflection and semantic feedback
Representative studies include Reflexion~\cite{shinn2023reflexion}, Self-Refine~\cite{madaan2023self}, GEPA~\cite{agrawal2025gepa}, SE-Agent~\cite{lin2025se}, and ACE~\cite{zhang2025agentic}.
Similarly, TextGrad~\cite{yuksekgonul2024textgrad} and REVOLVE~\cite{zhang2024revolve} interpret semantic feedback as textual gradients to guide model refinement.

Recently, experience-driven evolution~\cite{tang2025agent,zhou2025memento,ouyang2025reasoningbank,cai2025training} has emerged, allowing LLM agents to accumulate and reuse past interaction trajectories.
AgentKB~\cite{tang2025agent} emphasizes cross-domain knowledge transfer through a knowledge base. 
Memento~\cite{zhou2025memento} stores raw reasoning trajectories for later retrieval, but lacks generalization and interpretive guidance for novel contexts.
ReasoningBank~\cite{ouyang2025reasoningbank} scales memory retrieval at inference time without introducing an explicit learning process.
TF-GRPO~\cite{cai2025training} simply mimics the GRPO~\cite{shao2024deepseekmath} procedure to compute group-relative semantic advantages, but its extensibility and generalization are constrained by the GRPO-style design.

Overall, these approaches do not establish a principled learning paradigm that formalizes how LLM agents can \emph{learn from experience} in a continual, gradient-free manner. 
Moreover, most are validated only on simple reasoning benchmarks, without extending to more challenging scientific domains such as chemical retrosynthesis or protein fitness prediction.

    \section{Conclusion}
In this paper, we propose \method{}, a new paradigm that empowers LLM agents to continuously evolve through learning from experience.
\method{} constructs an ever-evolving experience library by extensive explorations and semantic distillation of environmental interactions during the learning process, which subsequently augments the model's capabilities of reasoning and leveraging expert knowledge without gradient back-propagation or parameter updates.
Extensive experiments demonstrate that \method{} significantly enhances the performance of baselines across diverse challenging scientific tasks, including mathematical reasoning, chemical retrosynthesis, and biological protein fitness prediction.
Conclusively, our work reveals the necessity of a fundamental shift in learning paradigms to address real-world challenges, thereby illuminating a promising pathway toward artificial intelligence endowed with open-ended evolution.

\section*{Acknowledgments and Disclosure of Funding}
This work is supported by the National Key R\&D Program of China (2022ZD0160501).
    
    \clearpage
    \bibliographystyle{unsrtnat}
    \bibliography{refs}

    \clearpage
    \beginappendix

\section{Experimental Details}

\subsection{Biology}


\textbf{Protein Fitness Prediction} We apply \method{} on the task of Protein Fitness Prediction. Proteins play a fundamental role in sustaining cellular functions that underpin organismal survival, growth, and reproduction. The capacity of a protein to perform its biological function—typically referred to as protein fitness—is determined by its three-dimensional structure, which is ultimately encoded by its amino acid sequence. Recent advances in machine learning have enabled zero-shot protein fitness prediction by leveraging representations derived from protein sequences, structural information, and auxiliary biological signals such as multiple sequence alignments (MSAs). Despite the progress, the task remains far from being fully solved. First, the absolute correlation between predicted and experimentally validated fitness scores remains limited (e.g., approximately 0.52 for the current state-of-the-art on ProteinGym), not confidential enough when encountering newly-found data with mysterious fitness. Second, no existing model consistently outperforms all others across diverse protein families and functional categories, including activity, expression, binding affinity, stability, and organismal fitness.

\paragraph{Agentic Task Transformation.} Large language model (LLM) agents, without specialized pre-training on protein sequences or domain-specific fine-tuning, inherently lack detailed biochemical knowledge and cannot directly infer subtle mutational effects on protein function. However, these agents possess strong capabilities in reasoning, ranking, knowledge synthesis, and few-shot learning. This opens a new opportunity: rather than directly predicting fitness from sequence, agents can be tasked with meta-reasoning over predictions made by high-performing protein language models. Specifically, we reformulate protein fitness prediction as an agentic feature selection and fusion task, where the agent learns to identify, prioritize, and integrate salient features derived from existing regression-based models. Through this process, the agent proposes explicit, interpretable algorithms that capture the underlying relationships among predictive signals, transforming opaque model outputs into a more accurate and explainable fitness prediction framework tailored to each target protein.




\paragraph{Benchmark.} We conduct experiments using the ProteinGym benchmark, which serves both as the source for experience extraction and as the evaluation dataset. Protein targets included in this benchmark align with those defined in SaProt~\cite{su2023saprot}. To emulate realistic biological discovery scenarios where experimental data is scarce, we adopt an imbalanced data split for each protein: a small experience set consisting of only 100 sequences (on average 1.47\% of the available data), and a large out-of-distribution test set containing all remaining sequences. This design reflects practical constraints in protein engineering, where only a limited number of labeled mutations are known, while the vast majority of sequence space remains unexplored.

\paragraph{Setup.} Prediction performance is evaluated using the Spearman rank correlation between the ground truth DMS (deep mutational scanning) scores and the proxy metrics generated by our \method{} method. Baseline models are grouped into two main categories:

\textit{Zero-shot Protein Fitness Prediction Models.} These models represent the current state-of-the-art on the ProteinGym leader-board and provide diverse biological priors through different input modalities:
\begin{itemize}
    \item VespaG~\cite{marquet2024expert, marquet2022embeddings}: A top-performing single-sequence model leveraging PLM embeddings and evolutionary guidance from GEMME~\cite{laine2019gemme}.
    \item PoET~\cite{truong2023poet}: A homolog-based model that encodes multiple sequence alignments using a sequence-of-sequences transformer architecture.
    \item ProSST~\cite{li2024prosst}: A hybrid model incorporating structural quantization and disentangled attention mechanisms to jointly model sequence and structural representations.
    \item VenusREM~\cite{tan2025high}: The current state-of-the-art on the ProteinGym scoreboard, integrating logits from single-sequence, homology, and structural tokenization signals.
\end{itemize}
These models provide the input features that our agent uses to reason about and construct enhanced fitness metrics.

\textit{Native Large Language Models (LLMs).} To isolate the contribution of agentic reasoning, we evaluate several high-performance LLMs—such as GPT-OSS~\cite{agarwal2025gpt} and Claude-Sonnet-4.5 as direct baselines. These models are prompted to perform the same prediction task as our forward learning system but without employing experience validation, rejection sampling and refinement. This comparison highlights the added value of structured forward learning beyond raw language model capabilities.


\begin{table}[t!]
\centering
\caption{Ablation study results in ProteinGym benchmark of forward learning paradigm component techniques: \textbf{Experience Exploration}, \textbf{Experience Evolution}, \textbf{Assistance of Regression tools}. Where Fixed mean best one for all regression tools is used instead of each different regression tools for protein target selected by agent.}
\label{tab:proteingym_ablation_matrix}

\resizebox{\textwidth}{!}{%
\begin{tabular}{ccccc}
\toprule
\textbf{Model} & \textbf{Experience Exploration} & \textbf{Experience Evolution} & \textbf{Regression Tools} & \textbf{Spearman's $\rho$} \\
\midrule
\multirow{7}{*}{GPT-OSS}
& \xmark & \xmark & \xmark & 0.472 \\
& \xmark & \xmark & Fixed & 0.531 \\
& \cmark & \xmark & \xmark & 0.537 \\
& \xmark & \cmark & \xmark & 0.547 \\
& \cmark & \cmark & \xmark & 0.573 \\
& \cmark & \cmark & Fixed & 0.568 \\
& \cmark & \cmark & \cmark & \textbf{0.581} \\
\bottomrule
\end{tabular}
} 
\end{table}

\textbf{Ablation Study.} From Table \ref{tab:proteingym_ablation_matrix} we can learn that, three core parts of our \method{} paradigm contributed to performance enhancement: Experience Exploration, Experience Evolution, Regression Tools. Specificlly, Experience Exploration provide success and failure cases agent used to engage to, in the ProteinGym Scenario, common features included in most successful cases suggest its effect-ness in prediction final proxy metric. Experience Evolution enables agent to summarize past experience, update lessons learned recently with new cross-validation facts. Moreover, based on accurate selection of top importance features, regression tools could help agents to further adjust limited hyper-parameters, where deciding what kinds of regression works best still important for agents compared to the fix, one for all regression cases.


\end{document}